\documentclass[a4paper,twoside]{article}
\pdfoutput=1
\usepackage[round]{natbib}\renewcommand{\cite}{\citep}

\usepackage{amsmath}
\usepackage{amsthm}
\usepackage{amssymb}
\usepackage{color}
\usepackage[conditional,first,outline,light]{draftcopy}
\usepackage{fancyhdr}
\usepackage[final]{graphicx}
\usepackage{ifthen}
\usepackage[latin1]{inputenc}
\usepackage{listings}
\usepackage{subfig}
\usepackage{stmaryrd}
\usepackage{array}
\usepackage{url}
\usepackage{xspace}
\makeatletter
\gdef\myfncyrunjournal{Preprint}
\ifthenelse{\lengthtest{\overfullrule=5pt}}%
  {\newcommand{\myfncyfoot}{\RCSId}}
  {\newcommand{\myfncyfoot}{\myfncyrunjournal}}

\def\ext@dash#1#2#3#4#5#6#7{%
  \mathrel{\mathop{%
    \setbox\z@\hbox{#5\displaystyle}%
    \setbox\tw@\vbox{\m@th
      \hbox{$\scriptstyle\mkern#3mu{#6}\mkern#4mu$}%
      \hbox{$\scriptstyle\mkern#3mu{#7}\mkern#4mu$}%
      \copy\z@
    }%
    \hbox to\wd\tw@{\unhbox\z@}}%
  \limits
    \@ifnotempty{#7}{^{\if0#1\else\mkern#1mu\fi
                       #7\if0#2\else\mkern#2mu\fi}}%
    \@ifnotempty{#6}{_{\if0#1\else\mkern#1mu\fi
                       #6\if0#2\else\mkern#2mu\fi}}}%
}
\def\dashfill@#1#2#3#4{%
  $\m@th\thickmuskip0mu\thickmuskip\thickmuskip\thickmuskip\thinmuskip
   \relax#4\raisebox{-.225ex}{$#1$}\mkern-7mu%
   \cleaders\hbox{$#4\mkern-2mu#2\mkern-2mu$}\hfill
   \mkern-7mu#3$%
}
\def\vdashfill@{\dashfill@\vdash\relbar\relbar}
\def\vDashfill@{\dashfill@\vDash\Relbar\Relbar}
\newcommand{\xvdash}[2][]{\ext@dash 5300\vdashfill@{#1}{#2}}
\newcommand{\xvDash}[2][]{\ext@dash 3095\vDashfill@{#1}{#2}}

\newenvironment{@useless}{}{}
\def\provideenvironment{\@star@or@long\provide@environment}
\def\provide@environment#1{%
  \@ifundefined{#1}
    {\new@environment{#1}}%
    {\renew@environment{@useless}}%
}

\def \@notp #1{\if #1\@false \else \@true \fi}

\def \@true {TT}
\def \@false {FL}
\def \@setflag #1=#2{\edef #1{#2}}

\long\def \@emptyargp #1{\@empargp #1\@empargq\@mark}
\long\def \@empargp #1#2\@mark{%
  \ifx #1\@empargq \@true \else \@false \fi}
\def \@empargq {\@empargq}

\newcommand{\acmcategories}{\small\noindent{\itshape ACM categories: }}
\@setflag \@firstcategory = \@true
\providecommand{\category}[3]{%
  \if \@firstcategory
    \\[.5em]\acmcategories
    \@setflag \@firstcategory = \@false
  \else
    \unskip;\hspace{.75em}%
  \fi
  \@ifnextchar [{\@category{#1}{#2}{#3}}{\@category{#1}{#2}{#3}[]}}
\def \@category #1#2#3[#4]{%
  {\let \and = \relax
   #1 [\textit{#2}]%
   \if \@emptyargp{#4}%
     \if \@notp{\@emptyargp{#3}}: #3\fi
   \else
     :\space
     \if \@notp{\@emptyargp{#3}}#3---\fi
     \textrm{#4}%
   \fi
  }
}

\gdef\myfncyrunauthor\@author
\gdef\myfncyruntitle\@title
\makeatother

\providecommand{\copyrightspace}{}
\providecommand{\inst}[1]{}
\providecommand{\and}{ and }
\providecommand{\institute}[1]{\date{\begin{small}\renewcommand{\and}{\\[1em]} #1\end{small}}}

\providecommand{\authorrunning}[1]{\gdef\myfncyrunauthor{#1}}

\providecommand{\titlerunning}[1]{\gdef\myfncyruntitle{#1}}

\providecommand{\tocauthor}[1]{}
\providecommand{\toctitle}[1]{}
\providecommand{\index}[1]{}
\providecommand{\href}[2]{\texttt{#2}}
\providecommand{\url}[1]{\href{#1}{#1}}
\providecommand{\email}[1]{\href{mailto:#1}{\texttt{#1}}}
\providecommand{\ead}[2][]{\email{#2}}
\providecommand{\journal}[1]{\gdef\myfncyrunjournal{#1}}

\provideenvironment{frontmatter}
  {%
    \renewcommand{\author}[2][]{}
    \renewcommand{\ead}[2][]{}
    
  }
  {}
\providecommand{\corauth}[2][]{}
\providecommand{\corauthref}[1]{}
\provideenvironment{abstract}
  {\small\begin{quotation}\noindent{\itshape R\'esum\'e\xspace}}
  {\end{quotation}}
\provideenvironment{keyword}
  {\small\begin{quotation}\noindent{\itshape Mots clefs :\xspace}}
  {\end{quotation}}

\providecommand{\doi}[1]{\href{http://dx.doi.org/#1}{doi: #1}}

\theoremstyle{plain}
\newtheorem{mytheorem}{Theorem}
\newtheorem{mylemma}{Lemma}
\newtheorem{myproposition}{Proposition}
\newtheorem{mycorollary}{Corollary}
\theoremstyle{definition}
\newtheorem{mydefinition}{Definition}
\newtheorem{myexample}{Example}

\provideenvironment{example}
  {\stepcounter{mytheorem}\begin{myexample}}{\qed\end{myexample}}
\provideenvironment{definition}
  {\stepcounter{mytheorem}\begin{mydefinition}%
   \renewcommand{\labelenumi}{\it(\roman{enumi})}}
  {\qed\end{mydefinition}}
\provideenvironment{corollary}
  {\stepcounter{mytheorem}\begin{mycorollary}%
   \renewcommand{\labelenumi}{\it(\roman{enumi})}}
  {\end{mycorollary}}
\provideenvironment{lemma}
  {\stepcounter{mytheorem}\begin{mylemma}%
   \renewcommand{\labelenumi}{\it(\roman{enumi})}}
  {\end{mylemma}}
\provideenvironment{proposition}
  {\stepcounter{mytheorem}\begin{myproposition}%
   \renewcommand{\labelenumi}{\it(\roman{enumi})}}
  {\end{myproposition}}
\provideenvironment{theorem}
  {\begin{mytheorem}\renewcommand{\labelenumi}{\it(\roman{enumi})}}
  {\end{mytheorem}}
\providecommand{\qed}{\hfill\mbox{$\square$}}
\provideenvironment{proof}{\textit{Proof.}~}{\qed\par}

\providecommand{\e}{\varepsilon}

\providecommand{\maketitle}{}
\newcommand{\fncy}{%
  \makeatletter
  \renewcommand{\ps@plain}{%
    \renewcommand\@evenhead{}
    \let\@oddhead\@evenhead
    \renewcommand\@evenfoot{\hfil\normalfont\myfncyfoot\hfil}
    \let\@oddfoot\@evenfoot
  }
  \pagestyle{fancy}
  \fancyhead{}
  \fancyhead[RO,LE]{\thepage}
  \fancyhead[LO]{\myfncyruntitle}
  \fancyhead[RE]{\myfncyrunauthor}
  \fancyfoot{}
  \cfoot{\myfncyfoot}
  \makeatother
}

\newcommand{\mi}[1]{\mathit{#1}}

\renewcommand{\a}{\alpha}
\newcommand{\gb}{\beta}

\renewcommand{\e}{\varepsilon}

\newcommand{\g}{\gamma}

\newcommand{\s}{\sigma}

\newcommand{\F}{\mathcal{F}}
\newcommand{\sD}{\mathcal{D}}
\newcommand{\sN}{\mathcal{N}}
\newcommand{\C}[1]{\mathcal{C}}

\renewcommand{\C}{\mbox{$\mathcal{C}$}} 

\newcommand{\G}{\mathcal{G}}                  

\renewcommand{\P}[1][]{\xrightarrow{#1}} 

\makeatletter 
\newcommand{\Xrightarrow}[2][]{\ext@arrow 0359\Rightarrowfill@{#1}{#2}}
\makeatother
\newcommand{\D} [1][]{\mathrel{\Xrightarrow{#1}}}                 
\newcommand{\Ds}[1]{\mathrel{\Xrightarrow{}^{#1}}}                
\newcommand{\Da}{\Ds{\ast}}                                       

\newcommand{\tup}[1]{\langle#1\rangle}

\newcounter{grammar}

\newcommand{\gdf}[2][]{
  \ifthenelse{\equal{#1}{}}%
    {%
      \[#2\]%
    }%
    {%
      \[#2\refstepcounter{grammar}\label{gram:#1}\eqno{(\G_{\text{\ref{gram:#1}}})}\]%
    }%
}

\newlength{\myhspace}
\newlength{\myminipagewidth}
\newcommand{\condlabel}[1]{(#1)}
\newlength{\condlength}
  { \begin{list}{}%
      { %
        \settowidth{\condlength}{\condlabel{#1}}%
        \setlength{\labelwidth}{\condlength}%
        \setlength{\labelsep}{1em}%
        \setlength{\itemsep}{0.5em}%
        \setlength{\leftmargin}{\labelwidth}%
        \addtolength{\leftmargin}{\labelsep}%
          { \begin{list}{\labelenumi}%
              { \usecounter{enumi}%
                \setlength{\itemsep}{0pt}%
                \addtolength{\leftmargin}{-\condlength}}}%
          {\end{list}}%
        \renewenvironment{itemize}%
          { \begin{list}{}%
              { \setlength{\itemindent}{-1.7em}%
                \setlength{\itemsep}{0pt}}}%
          {\end{list}}%
      }%
  }%
  {\end{list}}
\newcommand{\fstruct}[1]{%
  \setlength{\arraycolsep}{1pt}%
  \renewcommand{\arraystretch}{.5}%
  \left[\begin{array}{>{\scriptscriptstyle}r@{\scriptscriptstyle\;:\;}>{\scriptscriptstyle}l}%
       #1%
    \end{array}\right]}

\fncy
\journal{}
\usepackage[final]{hyperref}

\titlerunning{Feature Unification in TAG Derivation Trees}
\authorrunning{S. Schmitz and J. Le Roux}

\usepackage[figuresright]{rotating}
\title{Feature Unification in TAG Derivation Trees}

\author{Sylvain Schmitz\qquad Joseph Le Roux}
\institute{LORIA, INRIA Nancy Grand Est, France\\LORIA, Nancy
  Universit\'e, France\\\email{Sylvain.Schmitz@loria.fr}\qquad \email{Joseph.LeRoux@loria.fr}}
\begin{document}
\maketitle
\begin{abstract}
  The derivation trees of a tree adjoining grammar provide a first
  insight into the sentence semantics, and are thus prime targets for
  generation systems.  We define a formalism, \emph{feature-based
    regular tree grammars}, and a translation from feature based tree
  adjoining grammars into this new formalism.  The translation
  preserves the derivation structures of the original grammar, and
  accounts for feature unification.
\end{abstract}

\section{Introduction}
\begin{figure}[bt]
  \begin{center}
   \subfloat[Derived tree.]{\label{fig:derived}%
     \includegraphics[width=.57\linewidth]{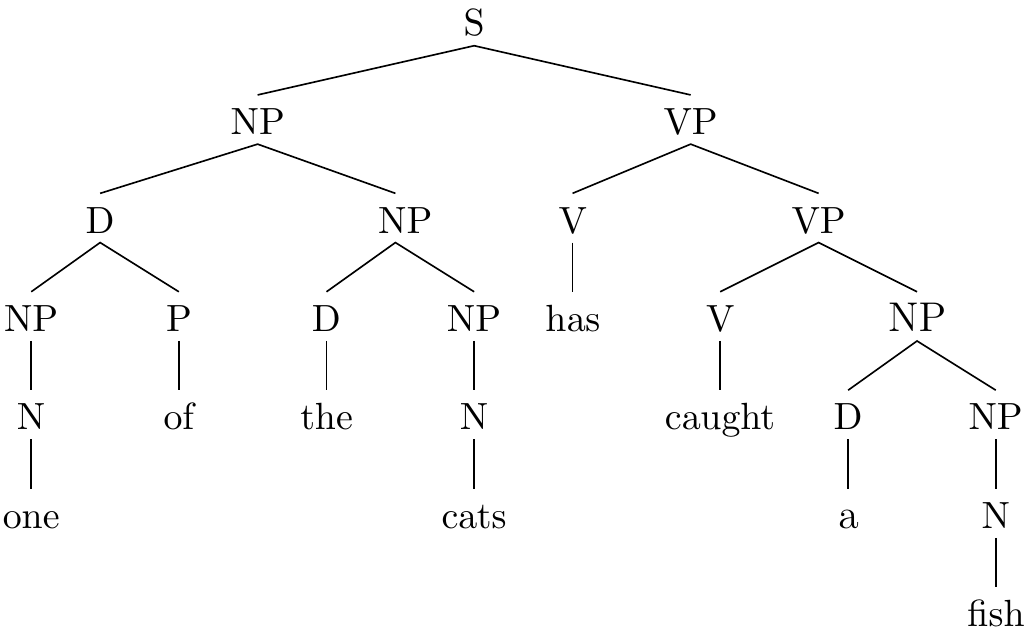}}
   \qquad
   \subfloat[Derivation tree.]{\label{fig:derivation}%
     \includegraphics[width=.32\linewidth]{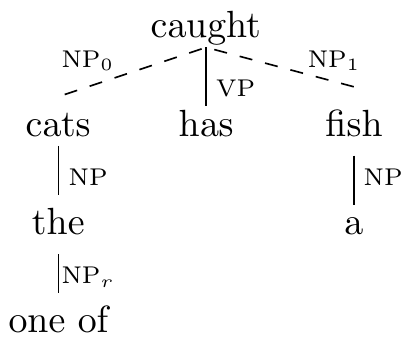}}
    \caption{Parse trees for ``One of the cats has
      caught a fish.'' using the grammar of \autoref{fig:gram}.}
  \end{center}
\end{figure}

Each sentence derivation in a tree adjoining grammar \cite[TAG]{tag}
results in two parse trees: a \emph{derived tree}
(\autoref{fig:derived}), that represents the phrase structure of the
sentence, and a \emph{derivation tree} (\autoref{fig:derivation}),
that records how the elementary trees of the grammar were
combined
.  Each type of parse tree is better suited for a different set of language
processing tasks: the derived tree is closely related to the lexical
elements of the sentence, and the derivation tree offers a first
insight into the sentence semantics \cite{mtt}.  Furthermore, the
derivation tree language of a TAG, being a regular tree language, is
much simpler to manipulate than the corresponding derived tree language.

Derivation trees are thus the cornerstone of several approaches to
sentence generation \cite{gendep,genplan}, that rely crucially on the
ease of encoding regular tree grammars, as dependency
grammars and planning problems respectively.  Derivation trees
also serve as intermediate representations from which both derived
trees (and thus the linear order information)
and semantics can be computed, e.g.\ with the abstract categorial grammars of
\citet{acgtag}, \citet{acgtagd}, and \citet{datalog}, or similarly with
the bimorphisms of \citet{bimorphisms}.
\begin{figure*}[t]
  \begin{center}
    \includegraphics[width=\textwidth]{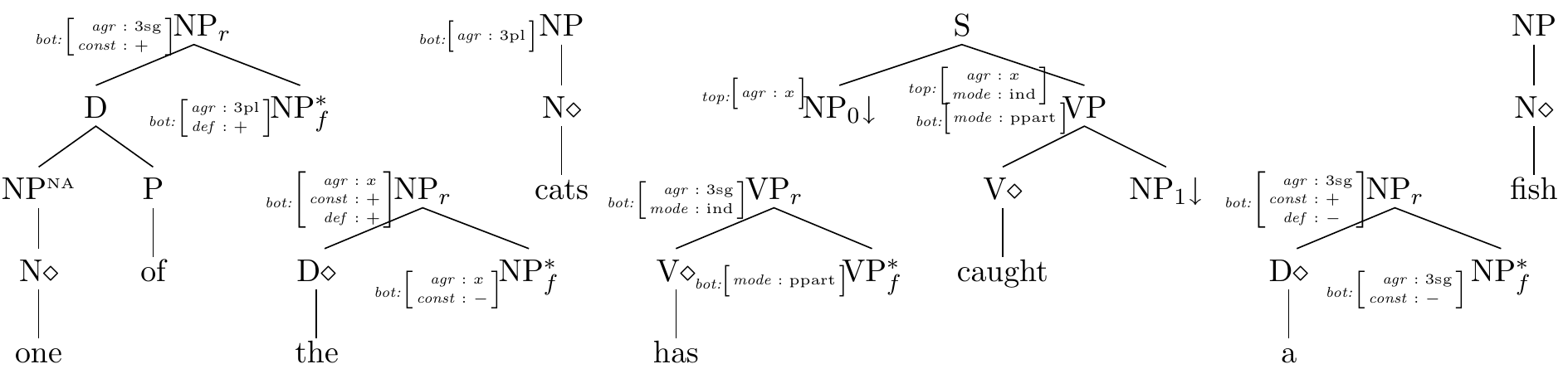}
    \caption{\label{fig:gram}A feature-based tree adjoining
      grammar. For the sake of clarity, we identify elementary trees with
      their anchors in our examples.}
  \end{center}
\end{figure*}

Nevertheless, these results do not directly apply to many real-world
grammars, which are expressed in a feature-based variant of TAGs
\cite{dfbtag}.  Each elementary tree node of these grammars carries
two feature structures that constrain the allowed substitution or
adjunction operations at this node (see for instance
\autoref{fig:gram}).  In theory, such structures are unproblematic,
because the possible feature values are drawn from finite domains, and
thus the number of grammar categories could be increased in order to
account for all the possible structures.  In practice, the sheer
number of structures precludes such a naive implementation: for
instance, the 50 features used in the XTAG English grammar
\cite{xtag} together define a domain containing more than $10^{19}$
different structures.  Furthermore, finiteness does not hold for some
grammars, for instance with the semantic features of
\citet{semfbtag}.

%
Ignoring feature structures typically results in massive
over-generation in derivation-centric systems.  We define
a formalism, \emph{feature-based regular tree grammars}, that produces
derivation trees that account for the feature structures found in a
tree adjoining grammar.  In more details,
\begin{itemize}
  \item we recall how to generate the derivation trees of a tree
    adjoining grammar through a regular tree grammar
    (\autoref{sec:rtg}), then
  \item we define feature-based regular tree grammars and present the
    translation from feature-based TAG (\autoref{sec:fbrtg}); finally,
  \item we provide an improved translation inspired by left corner
    transformations (\autoref{sec:lc}).%
\end{itemize}
%
%
%

We assume the reader is familiar with the theory of tree-adjoining
grammars \cite{tag}, regular tree grammars \cite{tata}, and
feature unification \cite{unif}.

\section{Regular Tree Grammars of Derivations}\label{sec:rtg}
In this section, we define an encoding of the set of derivation trees
of a tree adjoining grammar as the language of a regular tree
grammar (RTG).  Several encodings equivalent to regular tree grammars
have been described in the literature; we follow here the one of
\citet{acgtag}, but explicitly construct a regular tree grammar.

%
Formally, a tree adjoining grammar is a tuple $\tup{\Sigma,N,I,A,S}$
where $\Sigma$ is a terminal alphabet, $N$ is a nonterminal alphabet,
$I$ is a set of initial trees $\a$, $A$ is a set of auxiliary trees
$\gb$ and $S$ is a distinguished nonterminal from $N$. We note $\g_r$
the root node of the elementary tree $\g$ and $\gb_f$ the foot node of
the auxiliary tree $\gb$.  Let us denote by $\g_1,\ldots,\g_n$ the
\emph{active} nodes of an elementary tree $\g$, where a substitution
or an adjunction can be performed;\footnote{%
  We consider in particular that no adjunction can occur at a foot
  node.  We do not consider null adjunctions constraints on root nodes
  and feature structures on null adjoining nodes, which would rather
  obscure the presentation, and we do not treat other adjunction
  constraints either.
} we call $n$ the \emph{rank} of
$\g$, denoted by $\mathsf{rk}(\g)$.
We set $\g_1$ to be the root node of $\g$, i.e.\ $\g_1=\g_r$.  Finally,
$\mathsf{lab}(\g_i)$ denotes the label of node $\g_i$.

Each elementary tree $\g$ of the TAG will be converted into a single
rule \mbox{$X\P\g(Y_1,\ldots,Y_n)$} of our RTG, such that
$\mathsf{rk}(\g)=n$ and each of the $Y_i$ symbols represents the
possible adjunctions or substitutions of node $\g_i$.  We
introduce accordingly two duplicates \mbox{$N_A=\{X_A\mid X\in N\}$} and
\mbox{$N_S=\{X_S\mid X\in N\}$} of $N$, and a nonterminal labeling
function defined for any active node $\g_i$ with label
$\mathsf{lab}(\g_i)=X$ as
\begin{equation}
  \mathsf{nt}(\g_i)=\begin{cases}
  X_A& \text{if $\g_i$ is an adjunction site}\\
  X_S& \text{if $\g_i$ is a substitution site}
  \end{cases}
\end{equation}
The grammar rule corresponding to the elementary tree anchored by
``one of'' in \autoref{fig:gram} is then
\mbox{$\mathit{NP}_A\P\text{one of}(\mathit{NP}_A,D_A,P_A,N_A)$}, 
meaning that this tree adjoins into an $\mathit{NP}$ labeled node, and
expects adjunctions on its nodes $\mathit{NP}_r$, $D$, $P$, and $N$.
Given our set of elementary TAG trees, only the first one of these
four will be useful in a reduced RTG.

\begin{definition}
  The \emph{regular derivation tree grammar} $G=\tup{S_S,\sN,\F,R}$ of
  a TAG $\tup{\Sigma,N,I,A,S}$ is a RTG with axiom $S_S$, nonterminal
  alphabet $\sN=N_S\cup N_A$, terminal alphabet $\F=I\cup A\cup\{\e_A\}$
  with ranks $\mathsf{rk}(\g)$ for elementary trees $\g$ in
  $I\cup A$ and rank 0 for $\e_A$, and with set of rules
  \begin{align*}
   R =\;&\{X_S\P\a(\mathsf{nt}(\a_1),\dots,\mathsf{nt}(\a_n))
       \mid \a\in I, n=\mathsf{rk}(\a),X=\mathsf{lab}(\a_r)\}\\
  \cup\;&\{X_A\P\gb(\mathsf{nt}(\gb_1),\dots,\mathsf{nt}(\gb_n))
       \mid \gb\in A,n=\mathsf{rk}(\gb),X=\mathsf{lab}(\gb_r)\}\\
  \cup\;&\{X_A\P\e_A\mid X_A\in N_A\}\\[-2em]
  \end{align*}
\end{definition}

The $\e$-rules $X_A\P\e_A$ for each symbol $X_A$ account for
adjunction sites where no adjunction takes place.  The RTG has the
same size as the original TAG and the translation can be computed
in linear time.

\begin{example}\label{ex:rtg}
  The reduced regular tree grammar corresponding to
  the tree adjoining grammar of \autoref{fig:gram} is then:
  \par\vspace*{-1em}
  {\small\begin{align*}
      \langle&S_S,\{S_S,\mathit{VP}_S,\mathit{VP}_A,\mathit{NP}_S,\mathit{NP}_A\},\\
      &\{\text{one of},\text{the},\text{cats},\text{has},\text{caught},\text{a},\text{fish},\e_A\},\\
      &\{\:\:\:S_S\P\text{caught}(\mathit{NP}_S,\mathit{VP}_A,\mathit{NP}_S),\\
      &\;\mathit{NP}_S\P\text{cats}(\mathit{NP}_A),\\
      &\;\mathit{NP}_S\P\text{fish}(\mathit{NP}_A),\\
      &\;\mathit{NP}_A\P\text{the}(\mathit{NP}_A),\\
      &\;\mathit{NP}_A\P\text{a}(\mathit{NP}_A),\\
      &\;\mathit{NP}_A\P\text{one of}(\mathit{NP}_A),\\
      &\;\mathit{NP}_A\P\e_A,\\
      &\;\mathit{VP}_A\P\text{has}(\mathit{VP}_A),\\
      &\;\mathit{VP}_A\P\e_A\}\rangle
    \end{align*}}
  \par\vspace*{-2em}
\end{example}

Let us recall that the derivation relation induced by a regular tree
grammar $G=\tup{S_S,\sN,\F,R}$ relates terms\footnote{%
  The set of \emph{terms} over the alphabet $\F$ and the set of variables
  $\mathcal{X}$ is denoted by $T(\F,\mathcal{X})$;
  $T(\F,\emptyset)=T(\F)$ is the set of trees over $\F$.} %
of $T(\F,\mathcal{N})$, so that $t\D t'$ holds iff there exists a
context\footnote{%
    A \emph{context} $C$ is a term of
    $T(\F,\mathcal{X}\cup\{x\})$, $x\not\in\mathcal{X}$, which
    contains a single occurrence of $x$.  The term $C[t]$ for some
    term $t$ of $T(\F,\mathcal{X})$ is obtained by replacing this
    occurrence by $t$.} %
$C$ and a rule \mbox{$A\P a(B_1,\dots,B_n)$} such that $t=C[A]$ and
$t'=C[a(B_1,\dots,B_n)]$.  The language of the RTG is $L(G)=\{t\in
T(\F)\mid S_S\Da t\}$.
\begin{figure}[tb]
  \begin{center}
     \includegraphics[scale=.8]{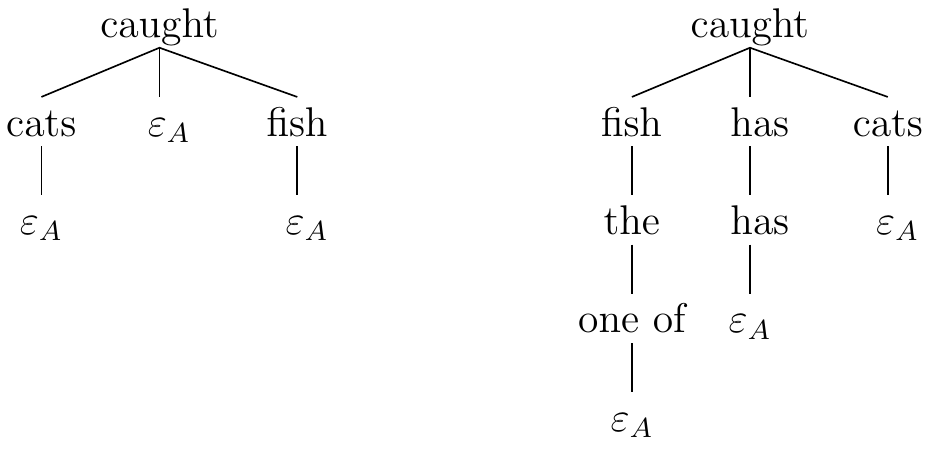}
    \caption{\label{fig:exrtg} Some trees generated by the regular
      tree grammar of \autoref{ex:rtg}.}
  \end{center}
\end{figure}

One can check that the grammar of \autoref{ex:rtg} generates
trees with a root labeled with ``caught'', and three subtrees, the
leftmost and rightmost of which labeled with ``cats'' or ``fish''
followed by an arbitrary long combination of nodes
labeled with ``one of'', ``a'' or ``the''.  The central subtree is an
arbitrary long combination of nodes labeled with ``has''.  Each branch
terminates with $\e_A$.  Two of these trees can be seen on
\autoref{fig:exrtg}. Our RTG generates the derivation trees of a
version of the original TAG expunged from its feature structures.

%

\section{Unification on TAG Derivation Trees}\label{sec:fbrtg}

\subsection{Feature-based Regular Tree Grammars}\label{sub:fbrtg}
%
%
In order to extend the previous construction to feature-based TAGs,
our RTGs use combinations of rewrites and unifications---also dubbed
\emph{narrowings} \cite{nrwngsurv}---of terms with variables in
$\sN\times\sD$, where $\sN$ denotes the nonterminal alphabet and $\sD$ the
set of feature structures.\footnote{In order to differentiate TAG tree
  substitutions from term substitutions, we call the latter
  \emph{u-substitutions}.  Given two feature structures $d$ and $d'$
  in $\sD$, we denote by the u-substitution $\s=\mathsf{mgu}(d,d')$
  their \emph{most general unifier} if it exists.  We denote by $\top$
  the most general element of $\sD$, and by $\mathit{id}$ the
  identity.}

\begin{definition}
  A \emph{feature-based regular tree grammar} $\tup{S,\sN,\F,\sD,R}$
  comprises an axiom $S$, a set $\sN$ of nonterminal symbols that
  includes $S$, a ranked terminal alphabet $\F$, a set $\sD$ of
  feature structures, and a set $R$ of rules of form $(A,d)\P
  a((B_1,d'_1),\dots,(B_n,d'_n))$, where $A,B_1,\dots,B_n$ are
  nonterminals, $d,d'_1,\dots,d'_n$ are feature structures, and $a$ is
  a terminal with rank $n$.

  The \emph{derivation} relation $\D$ for a feature-based RTG
  $G=\tup{S,\sN,\F,\sD,R}$ relates pairs of terms
  from $T(\F,\sN\times\sD)$ and u-substitutions, such that
  $(s,e)\D (t,e')$ iff there exist a context
  $C$, a rule $(A,d)\P a((B_1,d'_1),\dots,(B_n,d'_n))$ in $R$ with
  fresh variables in the feature structures, a structure $d'$, and an
  u-substitution $\s$ verifying
  \begin{gather*}
    s=C[(A,d')],\:t=C[a((B_1,\s(d'_1)),\dots,(B_n,\s(d'_n)))],\\
    \s=\mathsf{mgu}(d,e(d'))\text{ and }e'=\s\circ e .
  \end{gather*}
  The \emph{language} of $G$ is
  \begin{equation*}
    L(G)=\{t\in T(\F)\mid\exists e, ((S,\top),\mathit{id})\Da (t,e)\} .
  \vspace*{-1em}\end{equation*}
\end{definition}
Features percolate hierarchically through the computation of the most
general unifier $\mathsf{mgu}$ at each derivation step, while the
global u-substitution $e$ acts as an environment that communicates
unification results between the branches of our terms.

Feature-based RTGs with a finite domain $\sD$ are equivalent to regular
tree grammars.  Unrestricted feature-based RTGs can encode Turing
machines just like unification grammars \citep{avlg}, and thus we can
reduce the halting problem on the empty input for Turing
machines to the emptiness problem for feature-based RTGs, which is
thereby undecidable.

\subsection{Encoding Feature-based TAGs}\label{sec:fbtag}
%



For each tree $\g$ with rank $n$, we now create a rule
$P\P\g(P_1,\ldots,P_n)$.  A right-hand side pair
$P_i=(\mathsf{nt}(\g_i),d'_i)$ stands for an active node $\g_i$ with
feature structure
$d'_i=\mathsf{feats}(\g_i)=\fstruct{\mathit{top}&\mathsf{top}(\g_i)\\\mathit{bot}&\mathsf{bot}(\g_i)}$,
where $\mathsf{top}(\g_i)$ and $\mathsf{bot}(\g_i)$ denote
respectively the top and bottom feature structures of $\g_i$.

The left-hand side
pair $P=(A,d)$ carries the \emph{interface} $d=\mathsf{in}(\g)$ of
$\g$ with the rest of the grammar, such that $d$ percolates the root
$\mi{top}$ feature, and the foot $\mi{bot}$ feature for auxiliary
trees.  Formally, for each initial tree $\a$ in $I$ and auxiliary tree
$\gb$ in $A$, using a fresh variable $t$, we define
\begin{align}
  \mathsf{in}(\a)&=\fstruct{\mathit{top}&t\\\mathit{top}&\mathsf{top}(\a_r)}\\
  \mathsf{in}(\gb)&=\fstruct{\mathit{top}&t\\\mathit{top}&\mathsf{top}(\gb_r)\\\mathit{bot}&\mathsf{bot}(\gb_f)}
\end{align}
The interface thus uses the top features of the root node of an
elementary tree, and we have to implement the fact that this top
structure is the same as the top structure of the variable that
embodies the root node in the rule right-hand side. With the same
variable $t$, we define accordingly:
\begin{align}
  \mathsf{feat}(\g_i)=&
  \begin{cases}
    \fstruct{\mathit{top}&t\\\mathit{bot}&\mathsf{bot}(\g_r)}
            & \text{if } \g_i = \g_r \\
    \fstruct{\mathit{top}&\mathsf{top}(\g_i)\\
             \mathit{bot}&\mathsf{bot}(\g_i)}
              &     \text{otherwise}\\
  \end{cases}
\end{align}
Finally, we add $\e$-rules
$(X_A,\fstruct{\mathit{top}&v\\\mathit{bot}&v})\P\e_A$ for each symbol
$X_A$ in order to account for adjunction sites where no adjunction
takes place.  Let us denote by $\mathsf{tr}(\g_i)$ the pair
$(\mathsf{nt}(\g_i),\mathsf{feats}(\g_i))$.

\begin{definition}
  The feature-based RTG $G=\tup{S_S, N_S\cup
    N_A,\mathcal{F},\sD,R}$ of a TAG
  $\tup{\Sigma,N,I,A,S}$ with feature structures in $\sD$ has terminal
  alphabet $\mathcal{F}=I\cup A\cup \{\e_A\}$ with
  respective ranks $\mathsf{rk}(\a)$, $\mathsf{rk}(\gb)$, and $0$, and
  set of rules
\begin{align*}
  R&=\{(X_S,\mathsf{in}(\a))\P\a(\mathsf{tr}(\a_1),\dots,\mathsf{tr}(\a_n))\mid \a\in I, n=\mathsf{rk}(\a),X=\mathsf{lab}(\a_r)\}\\
  &\cup\{(X_A,\mathsf{in}(\gb))\P\gb(\mathsf{tr}(\gb_1),\dots,\mathsf{tr}(\gb_n))\mid \gb\in A,
  n=\mathsf{rk}(\gb),X=\mathsf{lab}(\gb_r)\}\\
 &\cup\{X_A\fstruct{\mathit{top}&t\\\mathit{bot}&t}\P\e_A\mid X_A\in N_A\}\\[-3em]
\end{align*}
\end{definition}

\begin{example}\label{eq:fbrtg}
With the grammar of \autoref{fig:gram}, 
we obtain the following ruleset:
\par\vspace*{-1.5em}
{\small
  \begin{equation*}
  \begin{array}{r@{\;\P\;}l}S_S\top &\text{caught}\left(\mi{NP}_S
      \fstruct{\mathit{top}&
        \fstruct{
          \mathit{agr}&x
        }\\
      },\mi{VP}_A
      \fstruct{
        \mathit{top}&
        \fstruct{
          \mathit{agr}&x\\
          \mi{mode}&\text{ind}
        }\\
        \mathit{bot}&
        \fstruct{
          \mathit{mode}&\text{ppart}
        }
      },\mi{NP}_S\top
      \right)\\[.4em]
    \mi{NP}_S\fstruct{\mathit{top}&t}&\text{cats}\left(\mi{NP}_A\fstruct{\mathit{top}&t\\\mathit{bot}&\fstruct{\mathit{agr}&3pl}}\right)\\[.3em]
    \mi{NP}_S\fstruct{\mathit{top} & t}&
    \text{fish}(\mi{NP}_A\fstruct{\mathit{top} & t})\\[.3em]
    \mi{NP}_A\fstruct{
        \mathit{top}&t\\
        \mathit{bot}&
        \fstruct{
          \mathit{agr}&x\\
          \mathit{const}&-
        }
    }
    &\text{the}\left(\mi{NP}_A
      \fstruct{
      \mathit{top}&t\\
      \mathit{bot}&
      \fstruct{
        \mathit{agr}&x\\
        \mathit{const}&+\\
        \mathit{def}&+
      }
    }
    \right)\\[.8em]
    \!\!\mi{NP}_A\fstruct{
        \mathit{top}&t\\
        \mathit{bot}&
        \fstruct{
          \mathit{agr}&\text{3sg}\\
          \mathit{const}&-
        }
    }
    &\text{a}\left(\mi{NP}_A
      \fstruct{
      \mathit{top}&t\\
      \mathit{bot}&
      \fstruct{
        \mathit{agr}&\text{3sg}\\
        \mathit{const}&+\\
        \mathit{def}&-
      }
    }
    \right)\\[.8em]
    \mi{NP}_A\fstruct{
        \mathit{top}&t\\
        \mathit{bot}&
        \fstruct{
          \mathit{agr}&\text{3pl}\\
          \mathit{def}&+\\
        }
    }
    &\text{one of}\left(\mi{NP}_A
      \fstruct{
      \mathit{top}&t\\
      \mathit{bot}&
      \fstruct{
        \mathit{agr}&\text{3sg}\\
        \mathit{const}&+\\
      }
    }
    \right)\\[.6em]
    \mi{NP}_A\fstruct{\mathit{top}&v\\\mathit{bot}&v}&\e_A\\
    \!\!\!\!\!\!\!\!\mi{VP}_A\fstruct{\mathit{top}&t\\\mathit{bot}&\fstruct{\mi{mode}&\text{ppart}}}&\text{has}\left(\mi{VP}_A\fstruct{\mathit{top}&t\\\mathit{bot}&\fstruct{\mi{agr}&\text{3sg}\\\mi{mode}&\text{ind}}}\right)\\[.4em]      
    \mi{VP}_A\fstruct{\mathit{top}&v\\\mathit{bot}&v}&\e_A\\[-1em]
   \end{array}
 \end{equation*}
}\end{example}

\begin{sidewaysfigure}[t]
  \begin{center}
    \includegraphics[width=\textwidth]{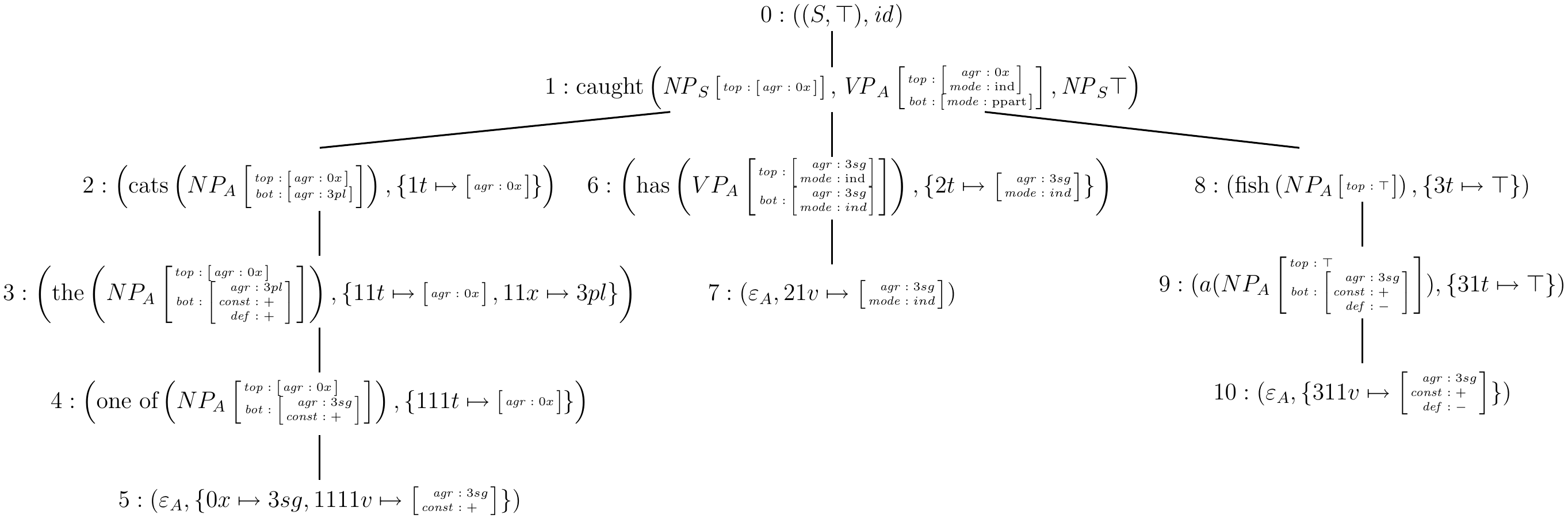}
    \caption{\label{fig:rtgderivation}A rewrite sequence in the
      feature-based RTG for the
      sentence ``One of the cats has caught a fish.''}
  \end{center}
\end{sidewaysfigure}

With the grammar of \autoref{eq:fbrtg}, one can generate the
derivation tree for ``One of the cats has caught a fish.'' This
derivation is presented in \autoref{fig:rtgderivation}.  Each node
of the tree consists of a label and of a pair $(t,e)$ where $t$ is a term
from $T(\F,\sN\times\sD)$ and $e$ is an environment.\footnote{%
  Actually,
  we only write the change in the environment at each point of the
  derivation.} %
In order to obtain fresh variables, we rename variables
from the RTG: we reuse the name of the variable in the grammar,
prefixed by the Gorn address of the node where the rewrite step takes
place.  Labels indicate the chronological order of the narrowings in
the derivation.

Labels in \autoref{fig:rtgderivation} suggest that this derivation
has been computed with a left to right strategy.  Of course, other
strategies would have led to the same result. The important thing to
notice here is that the crux of the derivation lies in the fifth
rewrite step, where the agreement between the subject and the verb is
realized. Substitutions sites are completely defined when all
adjunctions in the subtree have been performed. In the next section we
propose a different translation that overcomes this drawback.

\section{Left Corner Transformation}\label{sec:lc}
Derivations in the previous feature-based RTG are not very predictive:
the substitution of ``cats'' into ``caught'' in the derivation of
\autoref{fig:derivation} does not constrain the agreement feature of
``caught''.  This feature is only set at the final $\e$-rewrite step
after the adjunction of ``one of'', when the top and bottom features
are unified.  More generally, given a substitution site, we
cannot \textit{a priori} rule out the substitution of most initial
trees, because their root does usually not carry a top feature.

A solution to this issue is to compute the derivations in a
transformed grammar, where we start with the $\e$-rewrite, apply the
root adjunctions in reverse order, and end with the initial tree
substitution.  Since our encoding sets the root adjunct as the
leftmost child, this amounts to a selective left corner transformation
\cite{lc} of our RTG---an arguably simpler intuition than what we
could write for the corresponding transformation on derived trees.

\subsection{Transformed Regular Tree Grammars}
The transformation involves regular tree grammar rules of form
\mbox{$X_S\P\a(X_A,...)$} for substitutions, and
\mbox{$X_A\P\gb(X_A,...)$} and \mbox{$X_A\P\e_A$} for root adjunctions.
After a reversal of the recursion of root adjunctions, we will first
apply the $\e$ rewrite using a rule \mbox{$X_S\P\e_S(X)$} with rank 1
for $\e_S$, followed by the root adjunctions \mbox{$X\P\gb(X,...)$},
and finally the substitution itself \mbox{$X\P\a(...)$}, with a
decremented rank for initial trees. 

\begin{example}
  On the grammar of \autoref{fig:gram}, we obtain the rules:
  \par
{\small
  \begin{equation*}\label{lcrtg}
    \begin{array}{r@{\;\P\;}l}
      S_S & \text{caught}(\mathit{NP}_S,\mathit{VP}_A,\mathit{NP}_S)\\
      \mathit{NP}_S & \e_S(\mathit{NP})\\
      \mathit{NP}   & \text{cats}\\
      \mathit{NP}   & \text{fish}\\
      \mathit{NP}   & \text{the}(\mathit{NP})\\
      \mathit{NP}   & \text{one of}(\mathit{NP})\\
      \mathit{VP}_A & \text{has}(\text{VP}_A)\\
      \mathit{VP}_A & \e_A
    \end{array}\vspace*{-1em}
  \end{equation*}}
\end{example}

Adjunctions that do not occur on the root of an initial tree, like the
adjunction of ``has'' in our example, keep their original
translation using \mbox{$X_A\P\gb(X_A,...)$} and \mbox{$X_A\P\e_A$} rules.
We use the nonterminal symbols $X$ of the grammar for root adjunctions and
initial trees, and we retain $X_S$ for the initial $\e_S$ rewrite on
substitution nodes. 

\begin{definition}\label{def:lcrtg}
  The \emph{left-corner transformed} RTG $G_{\text{lc}}=\tup{S_S,N\cup
    N_S\cup N_A,\mathcal{F}_{\text{lc}},R_{\text{lc}}}$ of a TAG
  $\tup{\Sigma,N,I,A,S}$ has terminal alphabet
  $\mathcal{F}_{\text{lc}}=I\cup A\cup \{\e_A,\e_S\}$ with respective ranks
  $\mathsf{rk}(\a)-1$, $\mathsf{rk}(\gb)$, $0$, and $1$, and set
  of rules
  \begin{align*}
  R_{\text{lc}} &= \{X_S\P\e_S(X)\mid X_S\in N_S\}\\
  &\cup\{X\P\a(\mathsf{nt}(\a_2),\dots,\mathsf{nt}(\a_n))\mid \a\in I,
  n=\mathsf{rk}(\a),X=\mathsf{lab}(\a_r)\}\\
  &\cup\{X\P\gb(X,\mathsf{nt}(\gb_2)\dots,\mathsf{nt}(\gb_n))\mid \gb\in A,
  n=\mathsf{rk}(\gb),X=\mathsf{lab}(\gb_r)\}\\
  &\cup\{X_A\P\gb(\mathsf{nt}(\gb_1),\dots,\mathsf{nt}(\gb_n))\mid \gb\in A,
  n=\mathsf{rk}(\gb),X=\mathsf{lab}(\gb_r)\}\\
  &\cup\{X_A\P\e_A\mid X_A\in N_A\}\\[-3em]
  \end{align*}
\end{definition}

Due to the duplicated rules for auxiliary trees, the size of the
left-corner transformed RTG of a TAG is doubled at worst.  In
practice, the reduced grammar witnesses a reasonable growth
(10\% on the French TAG grammar of \citet{semfrag}).

The transformation is easily reversed.  We define accordingly the
function $\mathsf{lc}^{\text{-}1}$ from $T(\F_\text{lc})$ to $T(\F)$:
\begin{align*}
  \mathsf{lc}^{\text{-}1}(\e_S(t))&=\mathsf{s}(t,\e_A)\\
  \mathsf{s}(\gb(t_1,t_2,...,t_n),t)&=\mathsf{s}(t_1,\gb(t,f_{\gb_2}(t_2),...,f_{\gb_n}(t_n)))\\
  \mathsf{s}(\a(t_1,...,t_n),t)&=\a(t,f_{\a_2}(t_1),...,f_{\a_{n+1}}(t_n))\\
  \mathsf{a}(\g(t_1,...,t_n))&=\g(f_{\g_1}(t_1),...,f_{\g_n}(t_n))\\
  f_{\g_i}(t)&=\!\begin{cases}\mathsf{a}(t)&\text{if $\g_i$
      is an adjunction site}\\\mathsf{lc}^{\text{-}1}(t)&\text{if $\g_i$
      is a substitution site}\end{cases}
\end{align*}
We can therefore generate a derivation tree in $L(G_{\text{lc}})$
and recover the derivation tree in $L(G)$ through $\mathsf{lc}^{\text{-}1}$.

\subsection{Features in the Transformed Grammar}

\begin{example}
Applying the same transformation on the feature-based regular tree
grammar, we obtain the following rules for the grammar of
\autoref{fig:gram}:
\par\vspace*{-1em}
{\small
  \begin{equation*}\label{ex:lcfbrtg}
  \begin{array}{r@{\;\P\;}l}S_S\top &\text{caught}\left(\mi{NP}_S
      \fstruct{\mathit{top}&
        \fstruct{
          \mathit{agr}&x
        }\\
      },\mi{VP}_A
      \fstruct{
        \mathit{top}&
        \fstruct{
          \mathit{agr}&x\\
          \mi{mode}&\text{ind}
        }\\
        \mathit{bot}&
        \fstruct{
          \mathit{mode}&\text{ppart}
        }
      },\mi{NP}_S\top
      \right)\\[.4em]
      \mi{NP}_S\fstruct{\mathit{top}&t}&\e_S\left(\mi{NP}\fstruct{\mathit{top}&t\\\mathit{bot}&t}\right)\\
      \mi{NP}\fstruct{\mathit{bot}&\fstruct{\mathit{agr}&\text{3pl}}}&\text{cats}\\
      \mi{NP}\top&\text{fish}\\
      \mi{NP}\fstruct{
        \mathit{top}&t\\
        \mathit{bot}&
        \fstruct{
          \mathit{agr}&x\\
          \mathit{const}&+\\
          \mathit{def}&+
        }\\
      }
      &\text{the}\left(\mi{NP}
      \fstruct{
        \mathit{top}&t\\
        \mathit{bot}&
        \fstruct{
          \mathit{agr}&x\\
          \mathit{const}&-
        }\\
      }
      \right)\\[.8em]
      \mi{NP}\fstruct{
        \mathit{top}&t\\
        \mathit{bot}&
        \fstruct{
          \mathit{agr}&\text{3sg}\\
          \mathit{const}&+\\
          \mathit{def}&-
        }\\
      }
      &\text{a}\left(\mi{NP}
      \fstruct{
        \mathit{top}&t\\
        \mathit{bot}&
        \fstruct{
          \mathit{agr}&\text{3sg}\\
          \mathit{const}&-
        }\\
      }
      \right)\\[.8em]
      \mi{NP}\fstruct{
        \mathit{top}&t\\
        \mathit{bot}&
        \fstruct{
          \mathit{agr}&\text{3sg}\\
          \mathit{const}&+\\
        }\\
      }
      &\text{one of}\left(\mi{NP}
      \fstruct{
        \mathit{top}&t\\
        \mathit{bot}&
        \fstruct{
          \mathit{agr}&\text{3pl}\\
          \mathit{def}&+\\
        }\\
      }
      \right)\\[.5em]
      \mi{VP}_A\fstruct{\mathit{top}&t\\\mathit{bot}&\fstruct{\mi{mode}&\text{ppart}}}&\text{has}\left(\mi{VP}_A\fstruct{\mathit{top}&t\\\mathit{bot}&\fstruct{\mi{agr}&\text{3sg}\\\mi{mode}&\text{ind}}}\right)\\[.4em]      
      \mi{VP}_A\fstruct{\mathit{top}&v\\\mathit{bot}&v}&\e_A
   \end{array}
 \end{equation*}
}
\par\vspace*{-1em}
\end{example}
Since we reversed the recursion of root adjunctions, the feature
structures on the left-hand side and on the root node of the
right-hand side of auxiliary rules are swapped in their transformed
counterparts (e.g.\ in the rule for ``one of'').

This version of a RTG for our example grammar is arguably much easier
to read than the one described in \autoref{eq:fbrtg}: a
derivation has to go through ``one of'' and ``the'' before adding
``cats'' as subject of ``caught''.

The formal translation of a TAG into a transformed feature-based RTG
requires the following variant $\mathsf{tr}_{\text{lc}}$ of the
$\mathsf{tr}$ function: for any auxiliary tree $\gb$ in $A$ and any
node $\g_i$ of an elementary tree $\g$ in $I\cup A$, and with $t$ a
fresh variable of $\sD$:
\begin{align}
  \mathsf{in}_{\text{lc}}(\gb)&=\fstruct{\mathit{top}&t\\\mathit{bot}&\mathsf{bot}(\gb_f)}\\
  \mathsf{feats}_{\text{lc}}(\g_i)&=\begin{cases}
    \fstruct{\mathit{top}&t\\\mathit{top}&\mathsf{top}(\g_r)\\\mathit{bot}&\mathsf{bot}(\g_r)} &
    \text{if $\g_i=\g_r$}\\
    \mathsf{feats}(\g_i) &
    \text{otherwise}
  \end{cases}\\
  \mathsf{tr}_{\text{lc}}(\g_i)&=(\mathsf{nt}(\g_i),\mathsf{feats}_{\text{lc}}(\g_i))
\end{align}

\begin{definition}
  The \emph{left-corner transformed} feature-based RTG
  $G_{\text{lc}}=\tup{S_S,N\cup N_S\cup
    N_A,\mathcal{F}_{\text{lc}},\sD,R_{\text{lc}}}$ of a TAG
  $\tup{\Sigma,N,I,A,S}$ with feature structures in $\sD$ has terminal
  alphabet $\mathcal{F}_{\text{lc}}=I\cup A\cup \{\e_A,\e_S\}$ with
  respective ranks $\mathsf{rk}(\a)-1$, $\mathsf{rk}(\gb)$, $0$, and
  $1$, and set of rules
\begin{multline*}
  R_{\text{lc}}=\{X_S\fstruct{\mathit{top}&t}\P\e_S(X\fstruct{\mathit{top}&t\\\mathit{bot}&t})\mid X_S\in N_S\}\\     
  \shoveleft{\phantom{R_{\text{lc}}}\cup\{(X,\mathsf{feats}(\a_1))\P\a(\mathsf{tr}_{\text{lc}}(\a_2),\dots,\mathsf{tr}_{\text{lc}}(\a_n))}\\\shoveright{\mid \a\in I,n=\mathsf{rk}(\a),X=\mathsf{lab}(\a_r)\}}\\
  \shoveleft{\phantom{R_{\text{lc}}}\cup\{(X,\mathsf{feats}_{\text{lc}}(\gb_1))\P\gb((X,\mathsf{in}_{\text{lc}}(\gb)),\mathsf{tr}_{\text{lc}}(\gb_2),\dots,\mathsf{tr}_{\text{lc}}(\gb_n))}\\\shoveright{\mid \gb\in A, n=\mathsf{rk}(\gb),X=\mathsf{lab}(\gb_r)\}}\\
  \shoveleft{\phantom{R_{\text{lc}}}\cup\{(X_A,\mathsf{in}(\gb))\P\gb(\mathsf{tr}(\gb_1),\mathsf{tr}_{\text{lc}}(\gb_2),\dots,\mathsf{tr}_{\text{lc}}(\gb_n))}\\\shoveright{\mid \gb\in A, n=\mathsf{rk}(\gb),X=\mathsf{lab}(\gb_r)\}}\\     
  \shoveleft{\phantom{R_{\text{lc}}}\cup\{X_A\fstruct{\mathit{top}&t\\\mathit{bot}&t}\P\e_A\mid X_A\in N_A\}}\\[-2.5em]
\end{multline*}
\end{definition}

Again, the translation can be computed in linear time, and results in
a grammar with at worst twice the size of the original TAG.

\section{Conclusion}\label{sec:concl}
%
We have introduced in this paper feature-based regular tree grammars
as an adequate representation for the derivation language of large
coverage TAG grammars.
Unlike the restricted unification computations on the derivation tree
considered before by \citet{semunif}, feature-based RTGs accurately
translate the full range of unification mechanisms employed in TAGs.
Moreover, left-corner transformed grammars make derivations more
predictable, thus avoiding some backtracking in top-down
generation.

Among the potential applications of our results, let us further mention
more accurate reachability computations between elementary trees, needed
for instance in order to check whether a TAG complies with the tree
insertion grammar \cite[TIG]{tig} or regular form \cite[RFTAG]{rftag}
conditions.  In fact, among the formal checks one might wish to
perform on grammars, many rely on the availability of reachability
relations.  

Let us finally note that we could consider the string language of a
TAG encoded as a feature-based RTG---in a parser for instance---, if
we extended the model with topological information, in the line of
\citet{rdg}.

\providecommand{\Culik}{\v{C}ulik}\providecommand{\vd}{d\kern-.13em\raise.24ex%
\hbox{'}}

\end{document}